\documentclass[final]{cvpr}

\usepackage{times}
\usepackage{epsfig}
\usepackage{graphicx}
\usepackage{amsmath}
\usepackage{amssymb}
\usepackage{caption2}
\usepackage{subfigure}
\usepackage{bm}
\usepackage{booktabs}
\usepackage{multirow}
\usepackage[normalem]{ulem}
\useunder{\uline}{\ul}{}

\usepackage[pagebackref=true,breaklinks=true,colorlinks,bookmarks=false]{hyperref}

\def\cvprPaperID{327} 
\def\confYear{CVPR 2021}

\begin{document}


\title{Image-to-image Translation via Hierarchical Style Disentanglement}

\author{Xinyang Li$^{1}$,
Shengchuan Zhang$^{1}$\thanks{~Corresponding Author.},
Jie Hu$^{1}$,
Liujuan Cao$^{1}$,
Xiaopeng Hong$^{2}$,
Xudong Mao$^{1}$, 
\\
Feiyue Huang$^{3}$,
Yongjian Wu$^{3}$,
Rongrong Ji$^{1}$
\\
$^{1}$Xiamen University,
$^{2}$Xi'an Jiaotong University,
$^{3}$Tencent Youtu Lab
\\
{\tt\small 
\{imlixinyang, hujie.cpp, xudonmao\}@gmail.com,
\{zsc\_2016, caoliujuan, rrji\}@xmu.edu.cn,}
\\
{\tt\small 
hongxiaopeng@mail.xjtu.edu.cn,
\{garyhuang,littlekenwu\}@tencent.com
}
}

%

\maketitle

\begin{abstract}
Recently, image-to-image translation has made significant progress in achieving both multi-label (\ie, translation conditioned on different labels) and multi-style (\ie, generation with diverse styles) tasks.
However, due to the unexplored independence and exclusiveness in the labels, existing endeavors are defeated by involving uncontrolled manipulations to the translation results.
In this paper, we propose Hierarchical Style Disentanglement (HiSD) to address this issue.
Specifically, we organize the labels into a hierarchical tree structure, in which independent tags, exclusive attributes, and disentangled styles are allocated from top to bottom.
Correspondingly, a new translation process is designed to adapt the above structure, in which the styles are identified for controllable translations.
Both qualitative and quantitative results on the CelebA-HQ dataset verify the ability of the proposed HiSD.
%
%
The code has been released at \url{https://github.com/imlixinyang/HiSD}.
\end{abstract}

\section{Introduction}

Recently, deep learning based methods have achieved promising results in image-to-image translation area.
Early works~\cite{zhu2017unpaired,yi2017dualgan, liu2017unsupervised,taigman2016unsupervised} learn a deterministic mapping between two domains, which give rise to two emergent issues:
translating the inputs conditioned on multiple labels, and generating diverse outputs with multiple styles.
The former is termed the multi-label task, and the latter is termed the multi-style (or multi-modal) task.
For the multi-label task,  methods~\cite{Choi2017StarGAN, z._he_attgan:_2019, ming_liu_stgan:_2019, po-wei_wu_relgan:_2019} combine the labels into the translator.
For the multi-style task, methods~\cite{huang2018multimodal, lee2018diverse, amjad_almahairi_augmented_2018, jun-yan_zhu_toward_2017} incorporate latent codes drawn from Gaussian noise into the translator.
Recent unified solutions for these tasks can be classified into two categories.
(\romannumeral1). Works~\cite{romero2019smit, wang2019sdit, yu2019multi, li2019attribute} learn the shared style by injecting the style code concatenated with the target labels into the generator.
The shared style code does not have an explicit effect on the source image without changed labels, which is shown in Figure~\ref{fig.1}(a). 
(\romannumeral2). StarGANv2~\cite{yunjey_choi_stargan_2020} learns the mixed style by using the target label to index the mapped style code.
It continues to use the hypothesis of StarGAN~\cite{Choi2017StarGAN} that an image domain is the set of images sharing the same labels. 
The translations frequently involve unnecessary manipulations like changing facial identity and affecting background, as shown in Figure~\ref{fig.1}(b).
In addition, they cannot independently learn the respective styles for bangs, glasses, and hair color.
These uncontrollable translations severely limit their practical use.

\begin{figure}[t]
    \centering
    \includegraphics[width=1\linewidth]{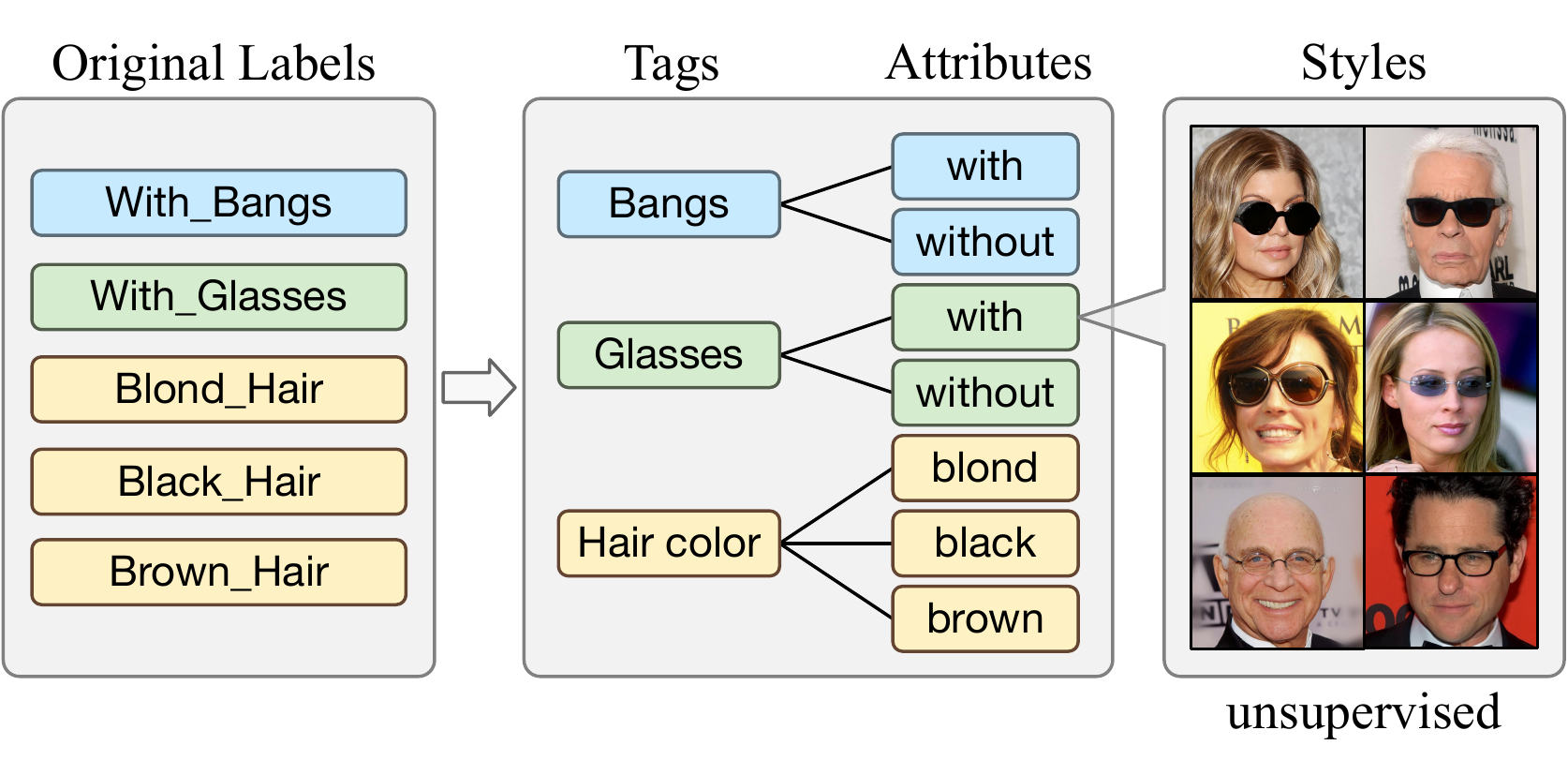}
    \caption{
    Hierarchical Style Disentanglement.
    The original labels are organized into independent tags and exclusive attributes. We aim to disentangle the styles to represent the clear manifestations in attributes, in an unsupervised way.
    }
    \label{fig.3}
\end{figure}

\begin{figure*}[t]
    \centering
    \includegraphics[width=1\textwidth]{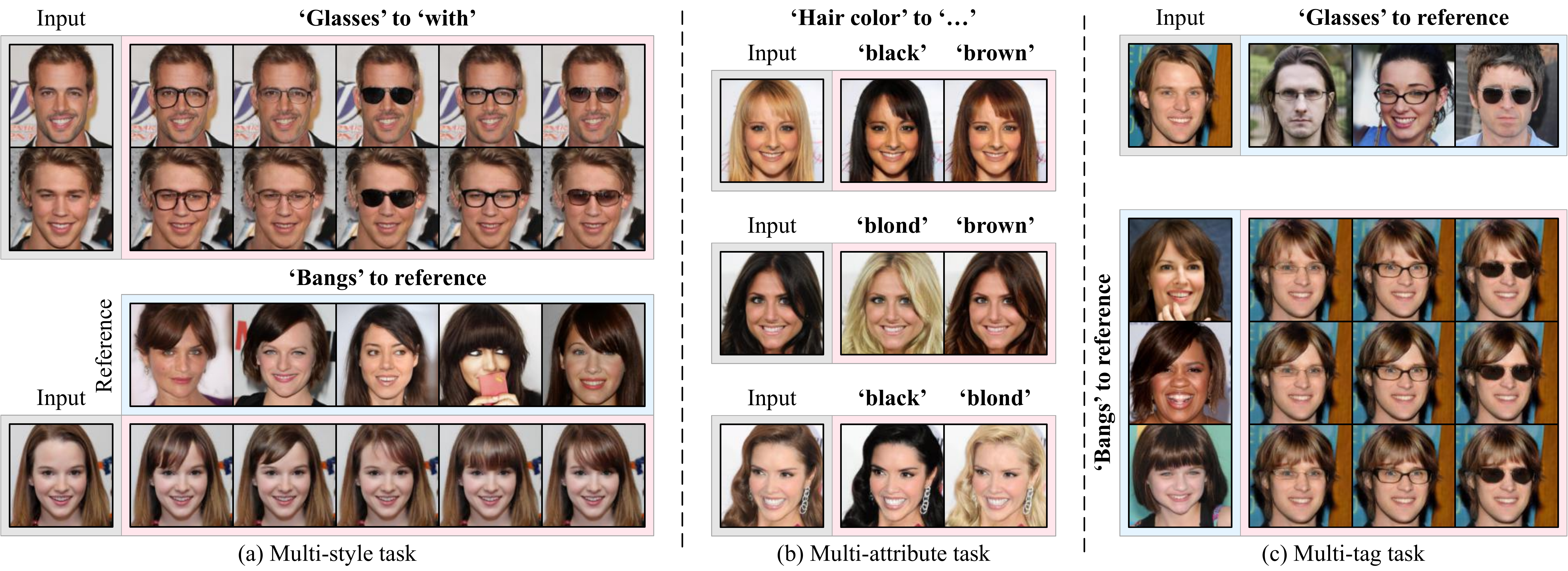}
    \caption{
    Selected results of our method on CelebA-HQ. 
    \textbf{(a)}. The multi-style task, which aims to generate diverse tag-relevant styles. The styles in our framework can be either generated by random latent codes or extracted from reference images.
    \textbf{(b)}. The multi-attribute task, which aims to translate images into multiple possible attributes.
    \textbf{(c)}. The multi-tag task, which aims to manipulate multiple tags of images simultaneously and independently. 
    } 
    \label{fig.2}
\end{figure*}

\begin{figure}[t]
    \centering
    \includegraphics[width=1\linewidth]{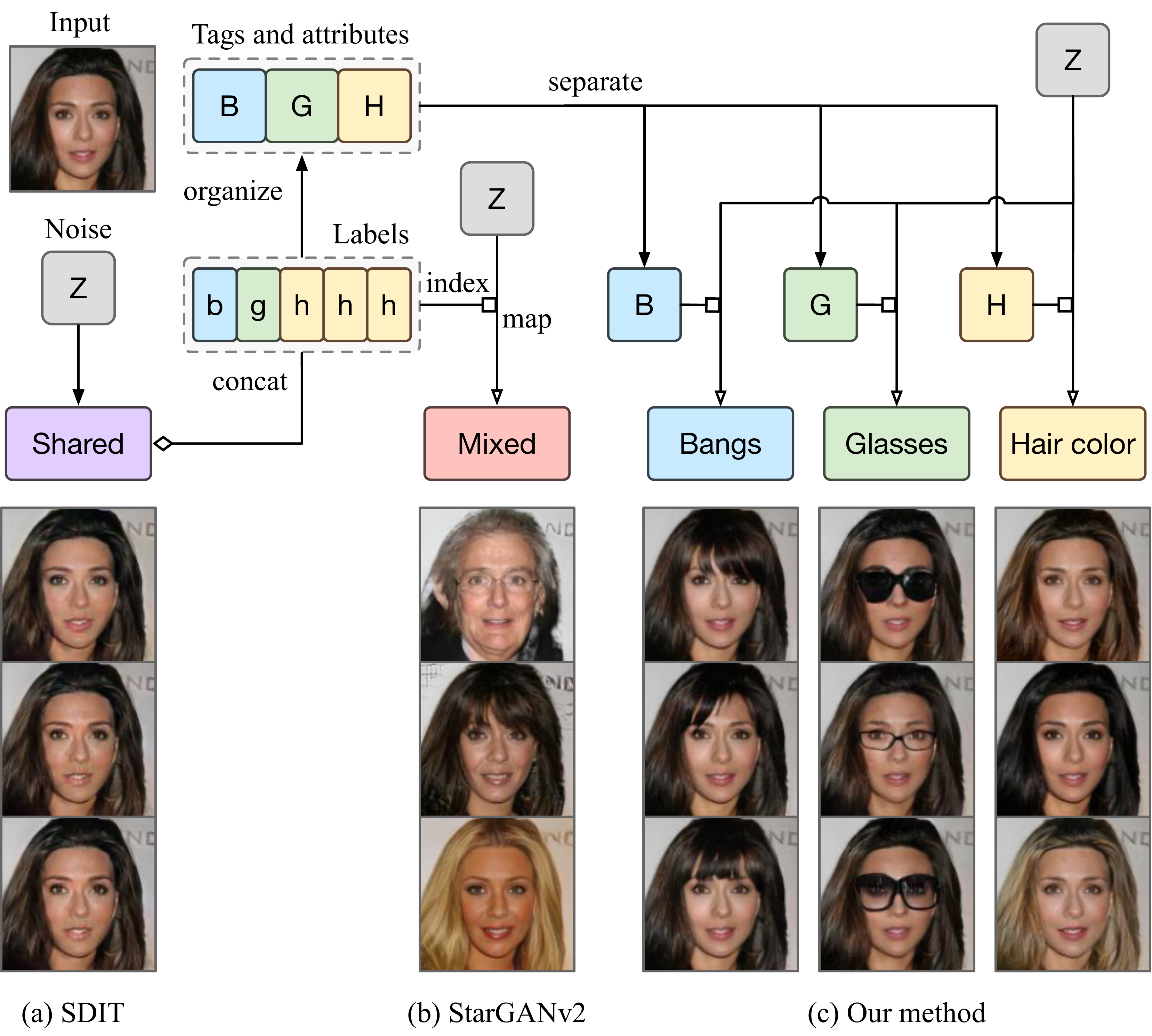}
    \caption{
    Comparison of different style codes. Our style codes are identified to the hierarchical structure.
    %
    }
    \label{fig.1}
\end{figure}

We propose a novel framework, called Hierarchical Style Disentanglement, to solve the above limitations.
We notice the general independence and exclusiveness among most label annotations.
For example, in CelebA, original binary labels `With\_Bangs' and `With\_Glasses' are independent, while `Blond\_Hair' and `Black\_Hair' are exclusive. 
Accordingly, as shown in Figure~\ref{fig.3}, we organize the original labels into a hierarchical structure, including independent tags and exclusive attributes.
The tags represent different accordance of attributes, and every image is relabeled to one of the attributes for each tag. 
For example, according to the tag `Glasses', the attribute of images can be either `with' or `without'.
Thus the multi-label issue is divided into two sub-tasks: multi-attribute task, which translates a tag to multiple possible attributes;
and multi-tag task, which manipulates multiple tags simultaneously.
However, the human-annotated attributes cannot represent the clear manifestations in images for tags.
In this paper, we take the clear manifestations in images for tags
as tag-relevant styles.
The tag-relevant styles, which are identified to the tags and attributes, provide a more controllable manner for diverse translations.
For example, the style for tag `Glasses' can disentangle different glasses, such as myopic glasses, sunglasses, and reading glasses in images, without supervised annotations.
We introduce different modules to generate, extract, and efficiently manipulate the disentangled tag-relevant styles.
In the cycle-translation path, we consistently optimize both generated and extracted styles to manipulate images realistically and accurately.
Through cycle consistency and style consistency, the generated and extracted styles are guaranteed to include the detailed manifestations for tags.
To guarantee the disentanglement, we introduce a local translator, which uses the attention mask to avoid the global manipulations;
and 
a tag-irrelevant conditional discriminator, which uses redundant labels in the annotations to prevent that these implicit conditions are manipulated by the translations.
In Figure~\ref{fig.2}, we show some selected results of our method on CelebA-HQ.

Our contributions include:
\begin{itemize}
  \item We propose HiSD to address the issues in recent multi-label and multi-style image-to-image translation methods by organizing the labels into a hierarchical structure, where independent tags, exclusive attributes, and disentangled styles are allocated from top to bottom.
  \item 
  To make the styles identified to the tags and attributes, we carefully redesign the modules, phases, and objectives.
  For unsupervised style disentanglement, we introduce two architectural improvements to avoid the global manipulations and implicit attributes to be manipulated during the translations.
  \item We conduct extensive experiments to prove the effectiveness of our model. 
\end{itemize}

\section{Related Works}

\noindent
\textbf{Generative Adversarial Networks.} GANs~\cite{goodfellow2014generative} have gained remarkable results.
After training, the generator is able to produce outputs which are similar to the real samples.
Recently, many works optimize the training stability of GANs~\cite{mao2017least,gulrajani2017improved,miyato2018spectral} or release the potential of GANs in different areas~\cite{reed2016generative,ledig2017photo,rippel2017real}.
Specifically, We use multi-task GANs~\cite{mescheder2018training,liu2019few} to make outputs satisfy the target attribute and conditional GANs~\cite{mirza2014conditional} to ensure that the tag-irrelevant conditions remain satisfying during translations. 

\noindent
\textbf{Image-to-image translation.}
Image-to-image translation has attracted increasing attentions since its widely practical use, such as colorization~\cite{zhang2016colorful}, super resolution~\cite{wu2017srpgan}, semantic synthesis~\cite{chen2017photographic} and domain adaption~\cite{hoffman2018cycada}.
Our framework focuses on the broad concept of image-to-image translation to jointly solve 
multi-label~\cite{Choi2017StarGAN, z._he_attgan:_2019, ming_liu_stgan:_2019, po-wei_wu_relgan:_2019} and multi-style~\cite{huang2018multimodal, lee2018diverse, amjad_almahairi_augmented_2018, jun-yan_zhu_toward_2017} issues, and overcomes the disadvantages of previous joint frameworks~\cite{romero2019smit, wang2019sdit, yu2019multi, li2019attribute, yunjey_choi_stargan_2020, xiao2017dna,zhou2017genegan, xiao2018elegant,jingtao_guo_mulgan:_2019}.

\noindent
\textbf{Label-specific style.}
Label-specific style learns the clear manifestation of the specific binary label, which is a special case of our tag-relevant style. 
However, it ignores the exclusiveness of some labels. 
Early works~\cite{xiao2017dna,zhou2017genegan,press2018emerging,chen2019homomorphic, benaim2019domain} use a single or multi-valued style code while recent methods~\cite{xiao2018elegant,jingtao_guo_mulgan:_2019} use the feature maps to represent the detailed manifestation.
However, they suffer from bad visual quality or unaligned images because of the pose-variant feature maps
and support only the reference-guided task.


\begin{figure*}[t!]
    \centering
    \includegraphics[width=\linewidth]{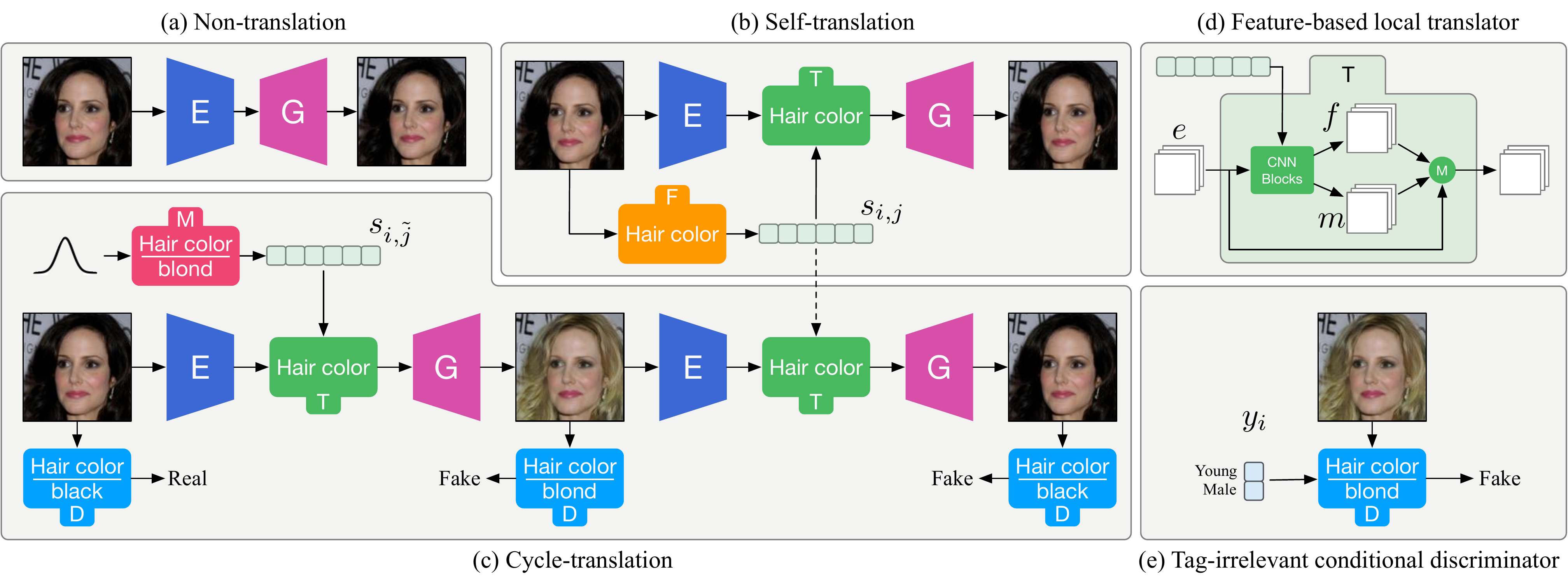}
    \caption{The training phases of our method include \textbf{(a)} a non-translation path, \textbf{(b)} a self-translation path, and \textbf{(c)} a cycle-translation path. We further avoid unnecessary manipulations by two essential architectural improvements: \textbf{(d)} a feature-based local translator and \textbf{(e)} a tag-irrelevant conditional discriminator.}
    \label{fig.5}
\end{figure*}



\section{Methods}

The organized labels can be defined as a hierarchical structure.
For a single image,
its attribute for tag $i \in \{1, 2, ..., N\}$ can be defined as
$
    j \in \{1, 2, ...,M_i\},
$
where $N$ is the number of tags and $M_i$ is the number of attributes for tag $i$.
However, the attribute $j$ cannot represent the detailed manifestation for the tag $i$ in images, which we called tag-relevant style $s_{i,j} \in \mathcal{S}_{i,j}$.
Similarly, we denote the image which has the attribute $j$ for tag $i$ by $x_{i,j} \in \mathcal{X}_{i,j}$.

\begin{figure}[t!]
    \centering
    \includegraphics[width=1\linewidth]{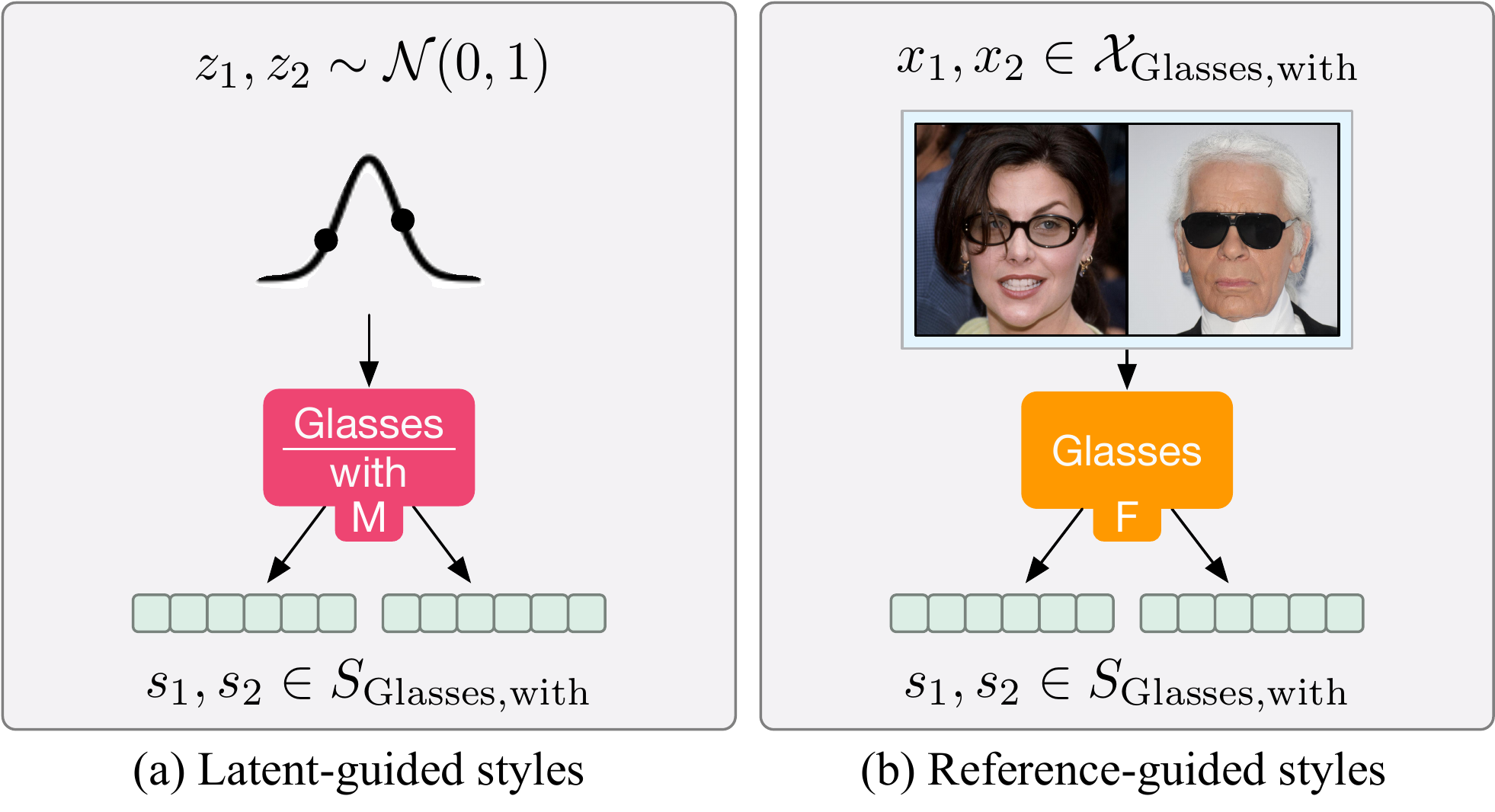}
    \caption{During test, the tag-relevant styles can be either \textbf{(a)} latent-guided (\ie generated by random latent codes) or \textbf{(b)} reference-guided (\ie extracted from reference images).}
    \label{fig.6l}
\end{figure}

Our framework aims to generate, extract, and manipulate the tag-relevant style in the image. 
Each purpose corresponds to a specific module.
There are two ways to get the style code for manipulation, as shown in Figure~\ref{fig.6l}.
For the latent-guided task, the style code is generated by the mapper module ($\mathbf{M}$).
Given a latent code $z \sim \mathcal{N}(0,1)$ and an attribute $j$ for tag $i$, $\mathbf{M}$ generates the style code $s_{i,j} = \mathbf{M}_{i,j}(z)$.
For the reference-guided task, the style code is extracted by the extractor module ($\mathbf{F}$).
Given an image $x_{i,j}$ and a tag $i$, $\mathbf{F}$ learns to extract the style code $s_{i,j} = \mathbf{F}_{i}(x_{i,j})$.
Then, the generated or extracted style is utilized to guide the manipulation.
However, it is inefficient if we apply manipulation directly to the image when manipulating multiple tags. 
Instead, denote a source image by $x$, we first convert it into its immediate feature by the encoder module ($\mathbf{E}$), which is given by $e = \mathbf{E}(x)$.
Then, to manipulate the feature, we introduce the translator module ($\mathbf{T}$).
Given a feature $e$ and a tag-relevant style code $s_{i,j}$ for tag $i$, $\mathbf{T}$ learns to manipulate the specific tag of the feature by $\mathbf{T}_i(e, s_{i,j})$. 
In each translation, the feature can go through the translators multiple times.
To get the translated image, we introduce the generator module ($\mathbf{G}$) to convert the translated feature $\tilde{e}$ into the image.
The translated image is given by $\tilde{x} = \mathbf{G}(\tilde{e})$.
We also introduce the discriminator module ($\mathbf{D}$) to determine whether an image, given the tag and attribute, is real or not.
Notably, for modules which need the tag or attribute as input,
we choose to use them to index the selection of specific layers of the modules rather than injecting them into the modules (\eg using the individual module $\mathbf{M}_{i,j}(\cdot)$ rather than a single $\mathbf{M}(\cdot, i, j)$ with tag $i$ and attribute $j$ as inputs).

\begin{figure}[t!]
    \centering
    \includegraphics[width=1\linewidth]{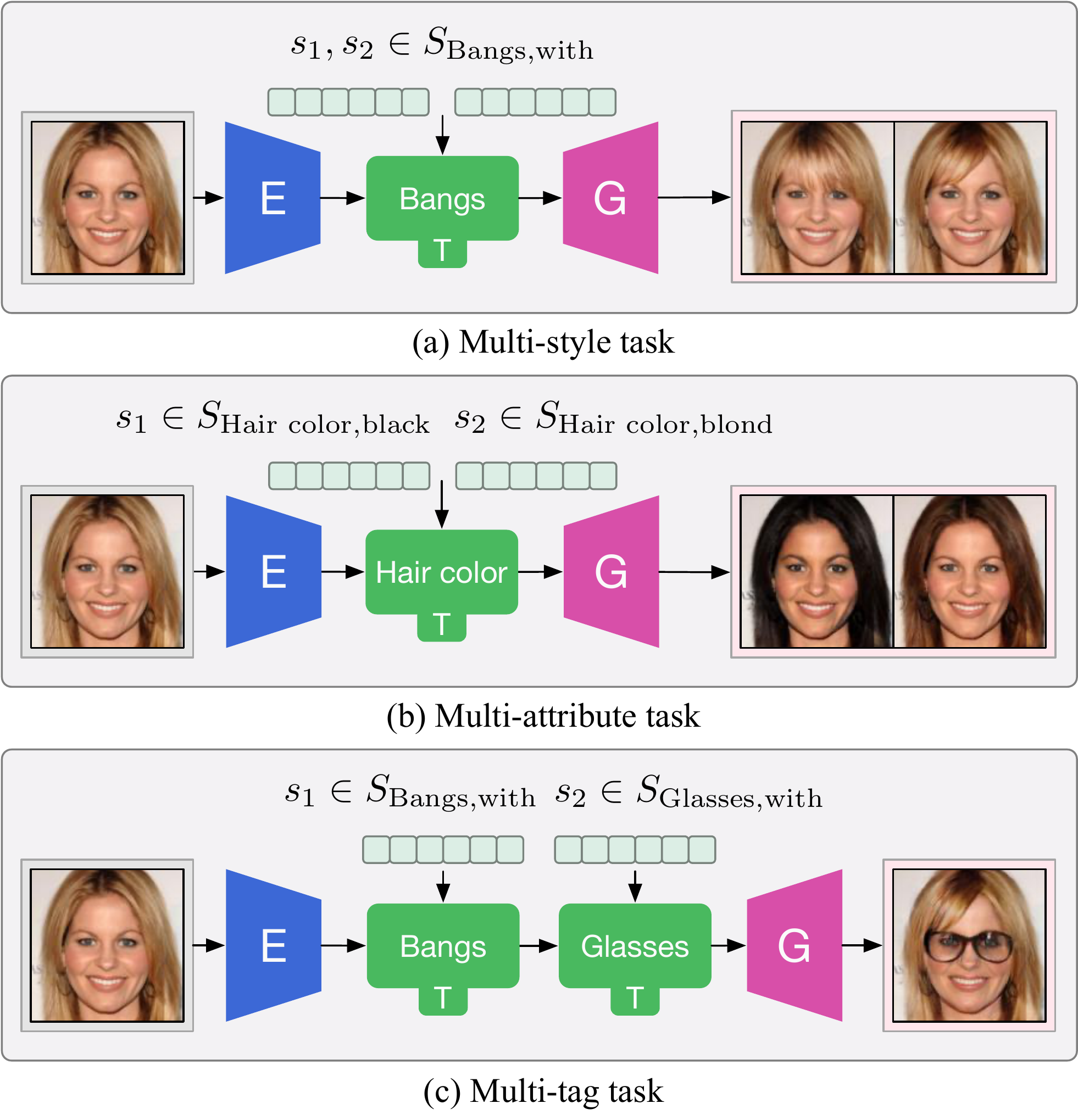}
    \caption{Test phases for \textbf{(a)} the multi-style task, \textbf{(b)} the multi-attribute task and \textbf{(c)} the multi-tag task. Note that the multi-tag task is not involved during training but achieved by the re-designed training phases and objectives indirectly.}
    \label{fig.6r}
\end{figure}

Formally, during the test, the source image $x$ is firstly encoded to its immediate feature 
\begin{equation}
    e_0 = e = \mathbf{E}(x).
\end{equation}
Second, to manipulate the source feature into target attributes $j_1, ..., j_l$ for multiple tags $i_1, ..., i_l$, respectively, where $l$ is the number of manipulated tags, 
we input the feature into the specific translator one by one, for $k = 1, ..., l$,
\begin{equation}
    e_k = \mathbf{T}_{i_k}(e_{k-1}, s_{i_k,j_k}),
\end{equation}
where $s_{i_k,j_k}$ can be either a latent-guided style $\mathbf{M}_{i_k,j_k}(z)$ or a reference-guided style $\mathbf{F}_{i_k}(x_{i_k,j_k})$. Finally, let $\tilde{e} = e_l$, the translated image is generated by
\begin{equation}
    \tilde{x} = \mathbf{G}(\tilde{e}).
\end{equation}
In particular, the test phases of the multi-style, multi-attribute, and multi-tag tasks are shown in Figure~\ref{fig.6r}.

\subsection{Training Phases}

To independently optimize the modules for different tags and attributes, 
we randomly sample a tag $i$, a source attribute $j$, and a target attribute $\tilde{j}$ in each iteration. As shown in Figure~\ref{fig.5}, given a source image $x_{i, j} \in \mathcal{X}_{i,j}$, the training phases include:

\noindent
\textbf{Non-translation path.}
We get the first reconstruction image $x_{i,j}' = \mathbf{G}(\mathbf{E}(x_{i,j}))$ in this path. 

\noindent
\textbf{Self-translation path.}
We get the second reconstruction image $x_{i,j}'' = \mathbf{G}(\mathbf{T}(\mathbf{E}(x_{i,j}), s_{i,j}))$, where $s_{i,j} = \mathbf{F}_i(x_{i,j})$ is the extracted tag-relevant style code of the source image. 


\noindent
\textbf{Cycle-translation path.}
In this path, we firstly generate the target tag-relevant style code $s_{i,\tilde{j}} = \mathbf{M}_{i,\tilde{j}}(z)$. Secondly, we render the generated style code $s_{i,\tilde{j}}$ into the feature of the source image $x_{i,j}$ and get the translated image $x_{i,\tilde{j}} = \mathbf{G}(\mathbf{T}(\mathbf{E}(x_{i,j}), s_{i,\tilde{j}}))$. Finally, the feature of the translated image $x_{i,\tilde{j}}$ and the original extracted style code $s_{i,j}$ are inputted into the translator and we get the third reconstruction image $x_{i,j}''' = \mathbf{G}(\mathbf{T}(\mathbf{E}(x_{i,\tilde{j}}), s_{i,j}))$.

\subsection{Training Objectives}


\noindent
\textbf{Adversarial objective.} The adversarial objective of our method encourages realistic manipulations for both generated and extracted styles, which is defined as
\begin{equation}
    \begin{split}
          \mathcal{L}_{adv}
          = 
          2\mathbb{E}_{i, j, x}[\log(\mathbf{D}_{i, j}(x_{i,j}))]&
           \\ 
          +\mathbb{E}_{i, j, x, \tilde{j}, z}[\log(1 - \mathbf{D}_{i, \tilde{j}}(x_{i,\tilde{j}}))]&
             \\
          +\mathbb{E}_{i, j, x, \tilde{j}, z}[\log(1 - \mathbf{D}_{i, j}(\tilde{x}_{i,j}'''))]&,
            \label{eq.4}
    \end{split}
\end{equation}
where $x_{i,\tilde{j}}$ is the translated image using the style code generated by the mapper $\mathbf{M}$,
and $\tilde{x}_{i,j}'''$ is the cycle-translated image using the style code extracted by the extractor $\mathbf{F}$. 
This objective not only encourages the mapper to accurately map the tag-specific attribute information into the generated tag-relevant style code, but also forces the extractor to extract the tag-specific attribute information from the image.

\noindent
\textbf{Reconstruction objective.}
All final outputs of the non-translation, self-translation, and cycle-translation paths are the reconstruction images of the source image. Thus, we apply a reconstruction objective to make the reconstruction images equal to the source image, which is
\begin{equation}
    \begin{split}
     \mathcal{L}_{rec}
          = 
          \mathbb{E}_{i, j, x}[\lVert 
          x_{i,j}' - x_{i,j}
          \rVert_1] & \\
          + \mathbb{E}_{i, j, x}[\lVert 
          x_{i,j}'' - x_{i,j}
          \rVert_1] & \\
          +  \mathbb{E}_{i, j, x, \tilde{j}, z}[\lVert 
          x_{i,j}''' - x_{i,j}
          \rVert_1] &.
    \end{split}
\end{equation}
Specifically, the first two terms encourage the consistency between the features whether through the translator $\mathbf{T}$ or not. 
They are significant for our framework to manipulate multiple tags during test, which is not directly involved during training.
The importance of these two terms has been proved in ModularGAN~\cite{bo_zhao_modular_2018}. 

The final term utilizes the cycle consistency to encourage the extracted tag-relevant style to be accurate, which needs models to extract the detailed manifestation of the source image and render it to manipulate the translated image, so that the cycle-translated image can be equal to the source one (\eg, the model needs to extract the specific style of glasses so that it can translate the non-glasses translated image back into the source image).


\noindent
\textbf{Style objective.} The extracted style code of the translated image is supposed to be equal to the generated style code. So we introduce the style objective
\begin{equation}
  \mathcal{L}_{sty}
  =
  \mathbb{E}_{i, j, x, \tilde{j}, z}[\lVert 
  \mathbf{F}_i(x_{i,\tilde{j}}) - s_{i,\tilde{j}}
  \rVert_1],
\end{equation}
which encourages the consistency between the generated and extracted styles~\cite{huang2018multimodal,lee2018diverse,wang2019sdit, li2019attribute,yu2019multi,yunjey_choi_stargan_2020}. 
On the one hand, it encourages the mapper $\mathbf{M}$ to generate the accurate tag-relevant style code, which can be equally extracted by the extractor $\mathbf{F}$ as well. 
On the other hand, it also encourages the translator $\mathbf{T}$ to fully utilize the style code and forces both generated and extracted styles to be camera-ready.

\noindent
\textbf{Full objective.} Finally, the optimization of the full objective function can be written as
\begin{equation}
  \min_{\mathbf{E},\mathbf{G},\mathbf{T},\mathbf{F},\mathbf{M}}\max_{\mathbf{D}}
  \mathcal{L}_{adv} + 
  \lambda_{rec} \mathcal{L}_{rec} +
  \lambda_{sty}\mathcal{L}_{sty},
\end{equation}
where $\lambda_{rec}$ and $\lambda_{sty}$ are hyper-parameters that control the relative importance of reconstruction and style objectives compared to the adversarial objective, respectively.
%
%
The full objective guarantees the style codes to catch the clear manifestations of different tags with unnecessary manipulations. We introduce two architectural improvements to avoid unnecessary manipulations and make the style codes further disentangled without extra objectives \cite{press2018emerging, benaim2019domain, hu2019information}.

\subsection{Feature-based Local Translator}

The translator in our framework affects the features rather than the images~\cite{youssef_alami_mejjati_unsupervised_2018,albert_pumarola_ganimation:_2018,hajar_emami_spa-gan:_2020, mokady2019mask}. To utilize the tag locality, we introduce a feature-based local translator. 
Denote the original feature by $e$, the translator outputs two folds of features $m$
and $f$ with the same size (\ie height, width, and channel) of $e$. Then the translated feature is given by
\begin{equation}
  \sigma(m) \cdot e + (1 - \sigma(m)) \cdot f,
\end{equation}
where $\sigma(\cdot)$ is the sigmoid function and $\sigma(m)$ is an attention mask. The attention mask in our translator is both spatial-wise and channel-wise. This design can avoid global manipulations like background and illumination during translations, with negligible additional calculation and no regularization objective.

\subsection{Tag-irrelevant Conditional Discriminator}

For different attributes, imbalanced phenomenons of implicit conditions are widespread in the real-world datasets.
In CelebA-HQ, there are 83.3\% male and 65.7\% aged in images with attribute `with' for tag `Glasses', while the percentages decrease to 36.0\% and 20.0\% correspondingly in images with attribute `without' for tag `Glasses'. 
The discriminator will force the translations to manipulate these implicit conditions.
We address this problem by injecting the tag-irrelevant conditions (\eg labels `Male' and `Young') into the discriminator.
Denote the tag-irrelevant conditions of the original image $x_{i,j}$ for tag $i$ by $y_i$, we replace Equation~\ref{eq.4} by
\begin{equation}
    \begin{split}
  \mathcal{L}_{adv}'
  = 
  2 \mathbb{E}_{i, j, x}[\log(\mathbf{D}_{i, j}(x_{i,j}, y_i))]
  & \\ 
  + \mathbb{E}_{i, j, x, \tilde{j}, z}[\log(1 - \mathbf{D}_{i, \tilde{j}}(x_{i,\tilde{j}}, y_i))]
  & \\
  +  \mathbb{E}_{i, j, x, \tilde{j}, z}[\log(1 - \mathbf{D}_{i, j}(x_{i,j}''', y_i))]&.
  \label{eq.9}
    \end{split}
\end{equation}
Consequently, the discriminator will notice the imbalanced phenomenons and encourage the translations to not manipulate the tag-irrelevant implicit conditions.

\section{Experiments}

\begin{figure*}[t!]
    \centering
    \includegraphics[width=1\linewidth]{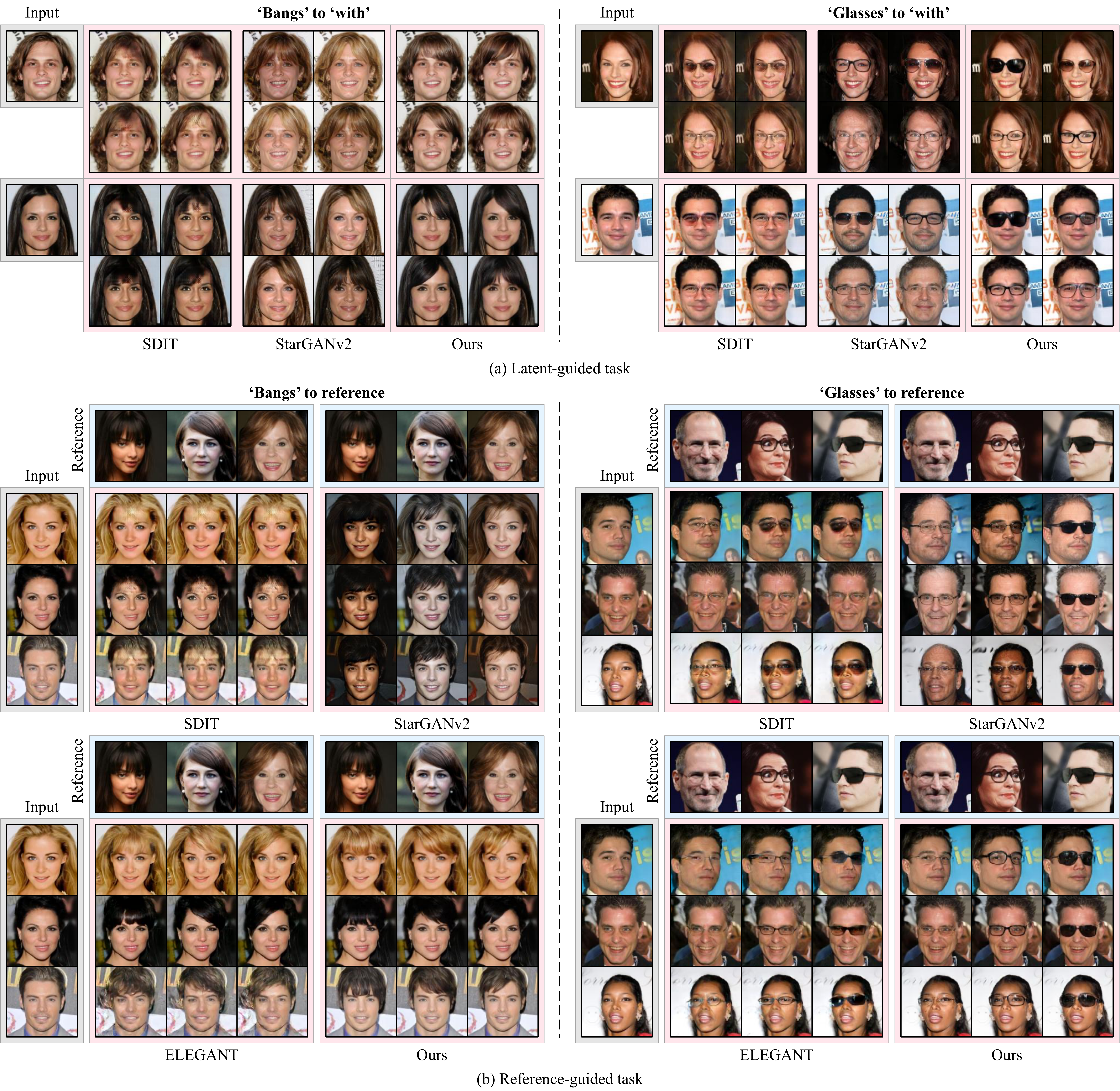}
    \caption{
    Qualitative results of the multi-style task, including \textbf{(a)} the latent-guided task and \textbf{(b)} the reference-guided task. Please zoom in for more details.
    }
    \label{fig.7}
\end{figure*}

In this section, we conduct a set of experiments to prove the effectiveness of our method.
All experiments are conducted using unseen images during training.

\noindent
\textbf{Dataset.} We choose CelebA-HQ~\cite{karras2018progressive} as our dataset, which contains 30,000 facial images with label annotations such as hair color, gender, and presence of glasses.
CelebA-HQ is more challenging than the original CelebA~\cite{liu2015faceattributes}.
The original labels `With\_Bangs', `With\_Glasses', `Blond\_Hair', `Black\_Hair', and `Brown\_Hair' in CelebA-HQ are organized into three tags `Hair color' (with attributes `blond', `black' and `brown'), `Bangs' (with attributes `with' and `without'), `Glasses' (with attributes `with' and `without') for experiments. For tag-irrelevant conditions, we choose two major labels, `Male' and `Young' to avoid the manipulation of gender and age.
We split the 30000 images in CelebA-HQ into 3000 images as the test set and 27000 as the training set.

\noindent
\textbf{Baselines.} We use SDIT~\cite{wang2019sdit} (with shared style), StarGANv2~\cite{yunjey_choi_stargan_2020} (with mixed style), and ELEGANT~\cite{xiao2018elegant} (with label-specific style) as our baselines. 
All models of baselines are trained using the implementations provided by their authors.
We train three independent models for StarGANv2 for each tag to avoid the exponential number of label domains.
For other implementation details, please refer to our supplemental material.


\subsection{Multi-style Task}

In this section, we evaluate our method on translating a tag of images into another attribute with diverse outputs.
We respectively manipulate tag `Bangs' to attribute `with' and tag `Glasses' to attribute `with' as our examples from two perspectives:
latent-guided task and reference-guided task.

\begin{table*}[t!]
\centering
\resizebox{0.7\textwidth}{!}{
\begin{tabular}{c||c|c|c|c|c|c|c|c|c}
\hline
\multirow{2}{*}{Method} & \multicolumn{3}{c|}{Realism(FID)}       & \multicolumn{3}{c|}{Diversity(User Study\%)} & \multicolumn{3}{c}{Disentanglement(FID)} \\ 
\cline{2-10} 
                  & L & R & G & L & R & G & L & R & G      \\ 
\cline{1-10}
SDIT~\cite{wang2019sdit} & 33.73 & 33.12 & 00.61 & 09.94 & 08.25 & 01.69 & 80.25 & 79.72 & 00.53 \\
\hline
StarGANv2~\cite{yunjey_choi_stargan_2020} & 26.04 & 25.49 & 00.55 & 29.18 & 26.63 & 02.55 & 90.08 & 78.03 & 12.05 \\
\hline
ELEGANT~\cite{xiao2018elegant} & - & 22.96 & - & - & 38.80 & - & - & 75.03 & -    \\
\hline
Ours & \textbf{21.37} & \textbf{21.49} & \textbf{00.12} & \textbf{50.00} & \textbf{48.97} & \textbf{01.03} & \textbf{71.85} & \textbf{71.48} & \textbf{00.37}     \\ \hline
\end{tabular}}
\caption{Quantitative results of the multi-style task. (\textbf{L}: latent-guided; \textbf{R}: reference-guided; \textbf{G}: gap between L and R.)
}
\label{tab.1}
\end{table*}

\begin{figure*}[t!]
    \centering
    \includegraphics[width=1\linewidth]{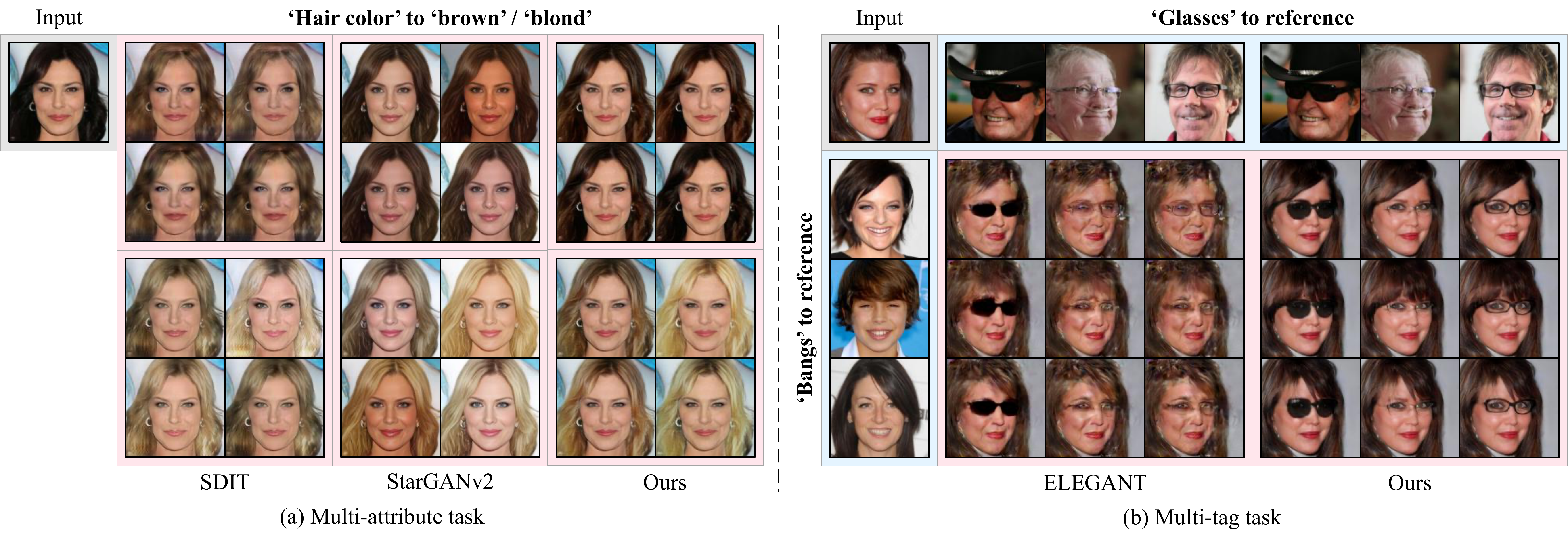}
    \caption{Qualitative results of \textbf{(a)} the multi-attribute task and \textbf{(b)} the multi-tag task.}
    \label{fig.9}
\end{figure*}

\noindent
\textbf{Latent-guided task.}
Figure~\ref{fig.7}(a) provides a qualitative comparison of the competing methods.
We cannot provide the results of ELEGANT because it only has the reference-guided capacity.
The visual quality and diversity of SDIT are limited.
StarGANv2 manipulates tag-irrelevant details (\eg changing the hair color when manipulating tag `Bangs' and translating the young female into an aged male when manipulating tag `Glasses').
%
Our method manipulates accurate tag-relevant styles for both tags, with high visual quality, satisfying diversity of the manipulated attributes, and maintaining tag-irrelevant details.

\noindent
\textbf{Reference-guided task.}
Figure~\ref{fig.7}(b) provides a qualitative comparison of the competing methods.
In this task, the styles are extracted from the reference images.
SDIT cannot effectively extract the styles. 
The expression of attributes comes from the change of the labels.
StarGANv2 extracts and manipulates many distinctive styles (\eg hair color, background, and facial identity).
ELEGANT succeeds in extracting and manipulating the accurate styles, but it generates visible artifacts in the last row of the left results.  
Another limitation is that the transfer of unaligned images is ineffective, because it uses pose-variant feature maps to represent the styles, while we use pose-invariant global style codes.
Our method transfers the styles for both tags accurately.

We also perform three quantitative comparisons between our method and baselines, including:

\noindent
\textbf{Realism.}
To qualitatively evaluate realism, we calculate the Frech\'{e}t inception distance (FID)~\cite{heusel2017gans}. 
For each test image without bangs, we translate it into images with bangs using 5 style codes,
which are generated by randomly sampled latent codes for the latent-guided task or extracted from random samples from images with bangs for the reference-guided task.
We then calculate the average FID between the translated images and real images with bangs.
Table~\ref{tab.1} shows the quantitative comparison of the competing methods.
Our method outperforms all baselines in terms of realism.

\begin{figure*}[t!]
    \centering
    \includegraphics[width=1\linewidth]{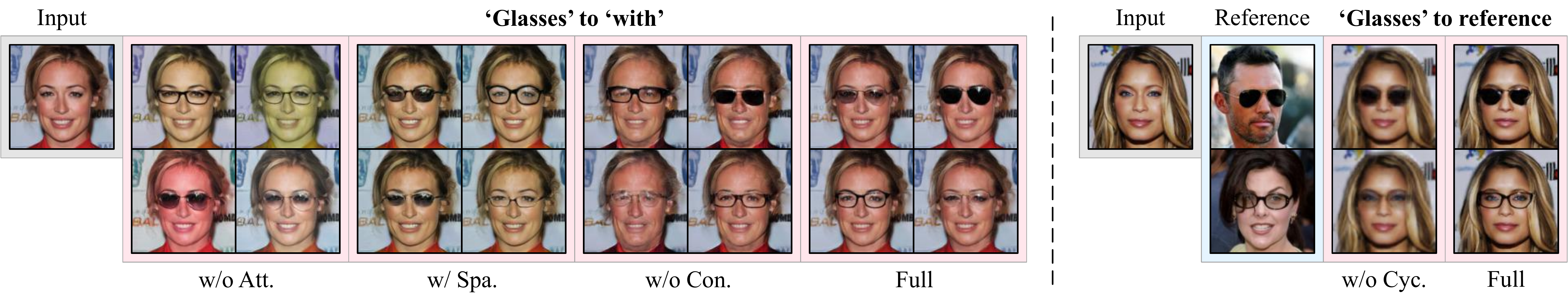}
    \caption{
    Qualitative results of the ablation study. 
    Results to the left of the dotted line are latent-guided while results to the right of the dotted line are reference-guided.
    }
    \label{fig.10}
\end{figure*}

\noindent
\textbf{Diversity.}
To qualitatively evaluate diversity, the widely used metric is the learned perceptual image patch similarity (LPIPS)~\cite{zhang2018unreasonable}. 
However, it encourages the models to manipulate the image as much as possible, which is not desired in practical use.
Hence, we choose to use user study for quantitative comparison of diversity. 
For each task, the percentage is given by asking users which diversity is more preferred, compared to the latent-guided results of our method.
%
The results in Table~\ref{tab.1} show that the baselines cannot obtain
a vote bigger than our method (\ie 50\%), which means that users prefer the diversity of our method more.
\noindent
\textbf{Disentanglement.}
To qualitatively evaluate the disentanglement of the tag-relevant styles,
we translate the young male images without bangs into 5 images with bangs and use the translated images and real young male images with bangs to calculate the average FID distance.
This metric challenges the methods to only transfer the tag-relevant styles and maintain the tag-irrelevant details.
If the model transfers tag-irrelevant details, such as changing the gender of the images, the FID between the translated images (some female images) and the real images (all male images) will increase.
The results show that the disentanglement of our method is superior to the baselines.

Moreover, we also report the capacity gap, which is the L1-norm between the values of the latent-guided task and reference-guided task.
The results show that our method gets the most balanced performance.

\subsection{Multi-label Task}

Multi-label task in previous methods is divided into two sub-tasks in our method: multi-attribute task and multi-tag task.
The former translates images into multiple possible attributes for one tag only. 
The latter manipulates multiple tags of images.

\noindent
\textbf{Multi-attribute task.}.
We translate the tag `Hair color' of images from `black' into two attributes `brown' and `blond' for example.
As shown in Figure~\ref{fig.9}(a),
SDIT gets worse visual quality than other methods, and StarGANv2 manipulates tag-irrelevant details (\eg skin color and illumination).
Our method accurately manipulates the disentangled tag-relevant styles, which satisfy both attributes for `Hair color'. The visual quality and diversity of our method are also desirable.

\noindent
\textbf{Multi-tag task.} 
An ideal method is supposed to independently manipulate each tag, which cannot be performed by SDIT and StarGAN due to their shared or mixed style code.
We manipulate two tags, `Bangs' and `Glasses', simultaneously. 
As shown in Figure~\ref{fig.9}(b), simultaneous translations of ELEGANT for multiple tags lead to interference (\eg changing the style of bangs while changing the style of glasses).
Furthermore, the results of our method demonstrate the disentanglement of our tag-relevant style codes.

\subsection{Ablation Study}

In this experiment, we evaluate several ablations of our method with different settings to investigate the effect of each additional component in our framework.
The qualitative and quantitative results are shown in Figure~\ref{fig.10} and Table~\ref{tab.2}, respectively.
The first ablation removes the feature-based local translator (w/o Att.).
The result shows that it cannot focus on the tag-relevant style (\eg changing the color and illumination of the images).
The second ablation uses the translator with only spatial-wise attention (w/ Spa.).
Although it can obtain tag-relevant diversity, the color distribution of the input image changes.
The third ablation removes the tag-irrelevant conditions for the discriminator (w/o Con.).
As expected, it is observed to change the gender and age of the input, but it achieves the best performance in terms of the realism due to the imbalanced phenomenon in the test images.
The last ablation removes the adversarial objective of the cycle-translation output (w/o Cyc.).
The results of this ablation in the latent-guided task are similar to the full model, but in the reference-guided task,
it is observed to output fuzzy results. 
The realism and disentanglement metrics of it in the reference-guided task are the worst among all ablations.
We suspect that the optimization with only the reconstruction objective of cycle-translation output is unclear for the framework, especially for small objects like glasses.
%
%
Our full model gets the best performance in qualitative results and sacrifices a bit of realism for disentanglement in quantitative results,
which proves the significance of each component in our framework.

\begin{table}[t!]
\centering
\resizebox{0.45\textwidth}{!}{
\begin{tabular}{c||c|c|c|c|c|c}
\hline
\multirow{2}{*}{Setting} & \multicolumn{3}{c|}{Realism(FID)}     & \multicolumn{3}{c}{Disentanglement(FID)} \\ 
\cline{2-7} 
         & L & R & G & L & R & G      \\ 
\cline{1-7}
w/o Att. & 22.30 & 21.83 & 00.47 & \underline{73.91} & \underline{71.50} & 02.41 \\
\hline
w/ Spa.  & 23.94 & 23.94 & \textbf{00.00} & 75.17 & 75.17 & \textbf{00.00} \\
\hline
w/o Con. & \textbf{20.63} & \textbf{20.66} & \underline{00.03} & 74.91 & 73.61 & 01.30 \\
\hline
w/o Cyc. & 22.31 & 28.39 & 06.08 & 74.18 & 80.34 & 06.16 \\
\hline
Full & \underline{21.37} & \underline{21.49} & 00.12 & \textbf{71.85} & \textbf{71.48} & \underline{00.37} \\ 
\hline
\end{tabular}}
\caption{Quantitative results of the ablation study.
         (\textbf{Bold}: best; \underline{Underline}: runner-up.)
         }
\label{tab.2}
\end{table}

\section{Conclusion}

In this paper, we propose the Hierarchical Style Disentanglement for image-to-image translation,
with scalability and controllable diversity. 
The key idea is to organize the labels into a hierarchical tree structure, which consists of independent tags, exclusive attributes, and unsupervised but disentangled styles. 
Extensive qualitative and quantitative experiments prove the effectiveness of our method.
Moreover, it is easy to adapt our method to joint training of multiple datasets~\cite{Choi2017StarGAN}, continuous learning of new tags~\cite{bo_zhao_modular_2018}, semi-supervised learning of semi-labeled dataset~\cite{li2019attribute}, and few-shot learning of tags with many attributes~\cite{liu2019few}.
We believe that our method is more suitable for practical use than the state-of-the-art image-to-image translation methods.

\clearpage

\section{Acknowledge}

This work is supported by the National Science Fund for Distinguished Young Scholars (No.62025603), the National Natural Science Foundation of China (No.U1705262, No. 62072386, No.62072387, No.62072389, No.62002305, No.61772443, No.61802324 and No.61702136) and Guangdong Basic and Applied Basic Research Foundation (No.2019B1515120049).

\renewcommand\thesection{\Alph{section}}

\section{Module Architecture}

\begin{figure*}[ht]
    \centering
    \includegraphics[width=1\linewidth]{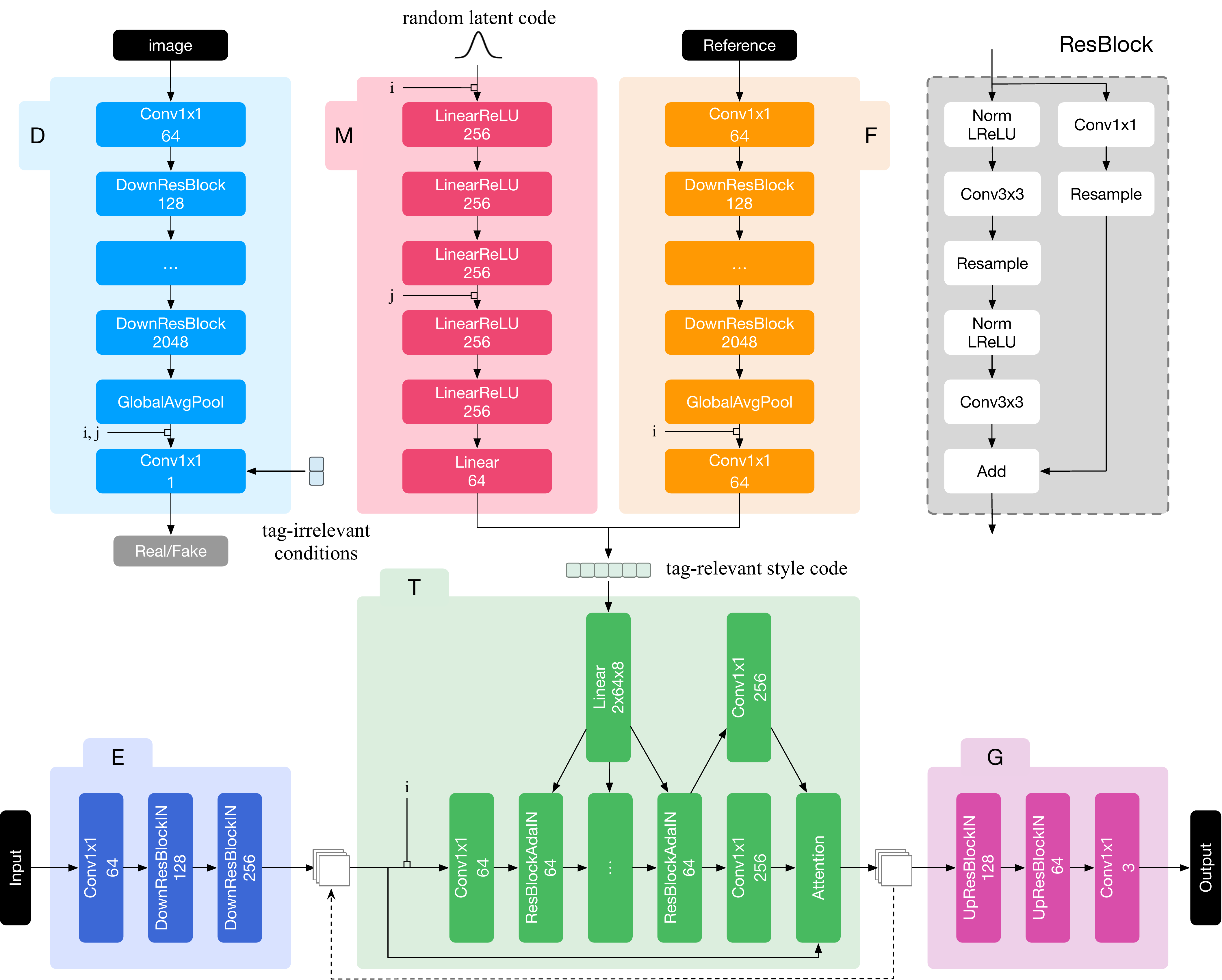}
    \caption{
     Architectural details of HiSD.
     To manipulate the input image, we first encode the image into its feature by $\mathbf{E}$.
     Then, the feature is manipulated by a single or multiple T.
     The manipulation is guided by the tag-relevant style code which can be either generated by $\mathbf{M}$ or extracted by $\mathbf{F}$.
     Finally, the output image is generated by $\mathbf{G}$.
     $\mathbf{D}$ is used to determine whether a image, given tag and attribute, is real or not.
     The details of ResBlocks (\ie DownResBlock, DownResBlockIN, ResBlockAdaIN, UpResBlockIN) are shown at the upper right corner.
    }
    \label{sm.1}
\end{figure*}

The architectural details of HiSD are shown in Figure~\ref{sm.1}. 
The mapper module ($\mathbf{M}$) consists of an MLP. The tag $i$ and attribute $j$ are used to index before the first layer and middle layer, respectively.
The extractor module ($\mathbf{F}$) consists of five downsampling blocks, of which inherit pre-activation residual units (ResBlock)~\cite{he2016identity}.
The tag $i$ is used to index before the last layer.
The encoder module ($\mathbf{E}$) consists of two downsampling blocks while the generator module ($\mathbf{G}$) consists of two upsampling blocks.
We use the Instance Normalization (IN)~\cite{ulyanov2016instance} in these two shared modules.
The translator module ($\mathbf{T}$) consists of eight immediate blocks with Adaptive Instance Normalization (AdaIN)~\cite{huang2017arbitrary} and decreased channel dimension.
The tag-relevant style code is injected into all AdaIN layers, providing scaling and shifting vectors through a linear layer.
The tag $i$ is used to index before the first layer.
The discriminator module ($\mathbf{D}$) uses the same architecture as $\mathbf{F}$ but uses both the tag $i$ and the attribute $j$ to index before the last layer.
For all residual units,
We use the Leaky ReLU (LReLU)~\cite{maas2013rectifier} as the activation function.
The Average Pooling and Nearest Neighbor Upsampling are used to resample the feature maps.

\section{Implementation Details}

The batch size is 8 and the model is trained for 200K iterations.
The images of CelebA-HQ are resized to 128$\times$128.
The training time is
around 40 hours on a single GTX 1080Ti GPU with our
implementation in PyTorch~\cite{paszke2017automatic}. 
To stabilize the training, we adopt the hinge version of adversarial loss~\cite{miyato2018spectral} with R1-regularization~\cite{mescheder2018training} using $\gamma$ = 1. We use the Adam~\cite{kingma2014adam} optimizer with $\beta_1$ = 0 and $\beta_2 = 0.99$. 
The learning rate is $0.0001$ except for the mapper, of which the learning rate is $0.000001$~\cite{karras2019style}.
We use the historical average version~\cite{karras2018progressive} of the intermediate modules for test where the update weight is 0.001.
We initialize the weights of all modules using He initialization~\cite{he2015delving} and set all biases to zero, except for the biases associated with the scaling vectors of AdaIN that are set to one.
For hyper-parameters, we easily set $\lambda_{rec}=1$ and $\lambda_{sty}=1$ in all experiments.

\section{Comparison without Cherry-picking}

\begin{figure*}[ht]
    \centering
    \includegraphics[width=1\linewidth]{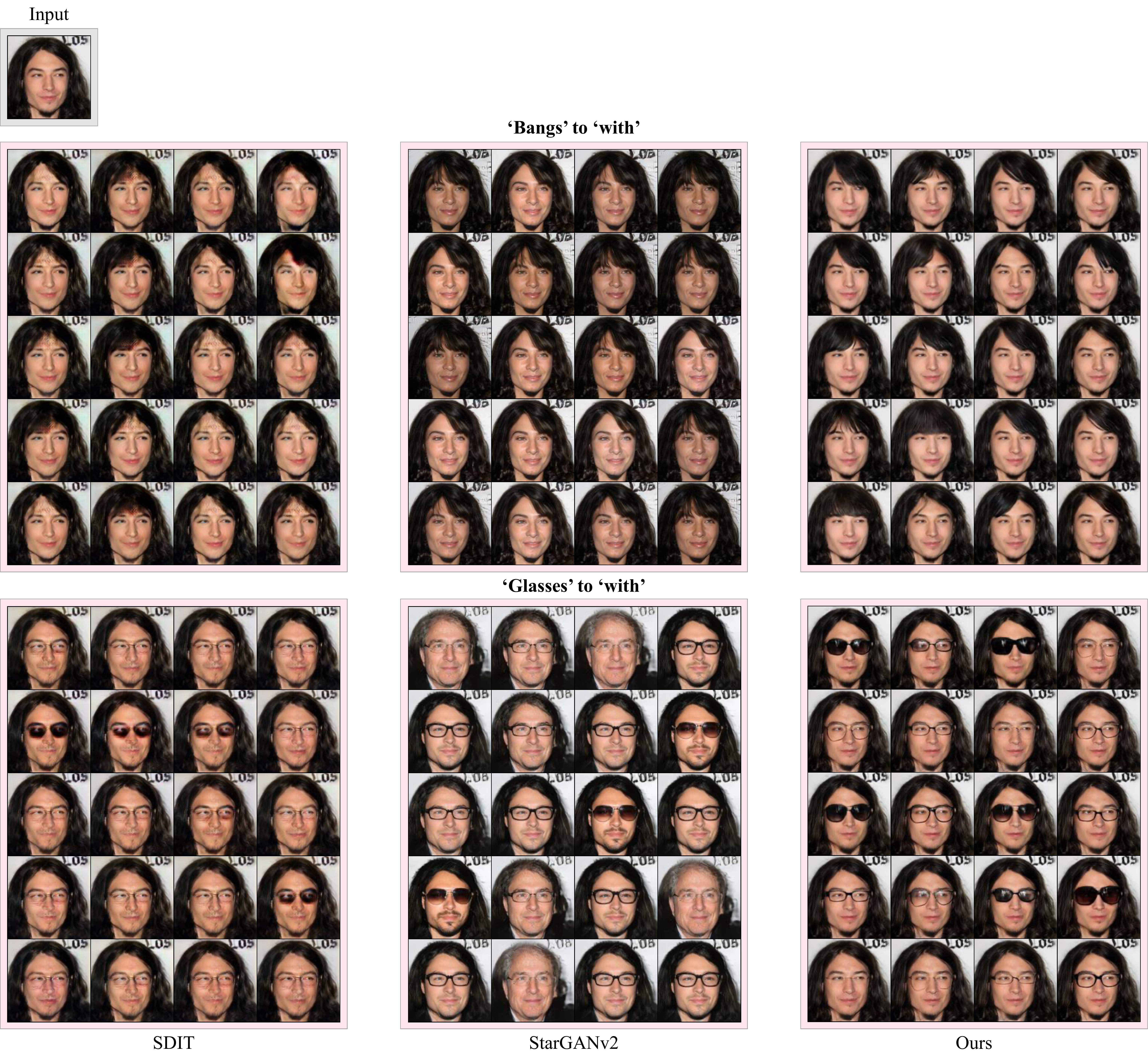}
    \caption{
     Additional qualitative results without cherry-picking  of the baselines and our method for the latent-guided multi-style task. We respectively manipulate the input image to attribute `with' for tag `Bangs' and `Glasses' by using 20 random latent codes, which are drawn from Gaussian distribution, to generate diverse outputs.
    }
    \label{sm.2-1}
\end{figure*}

\begin{figure*}[ht]
    \centering
    \includegraphics[width=1\linewidth]{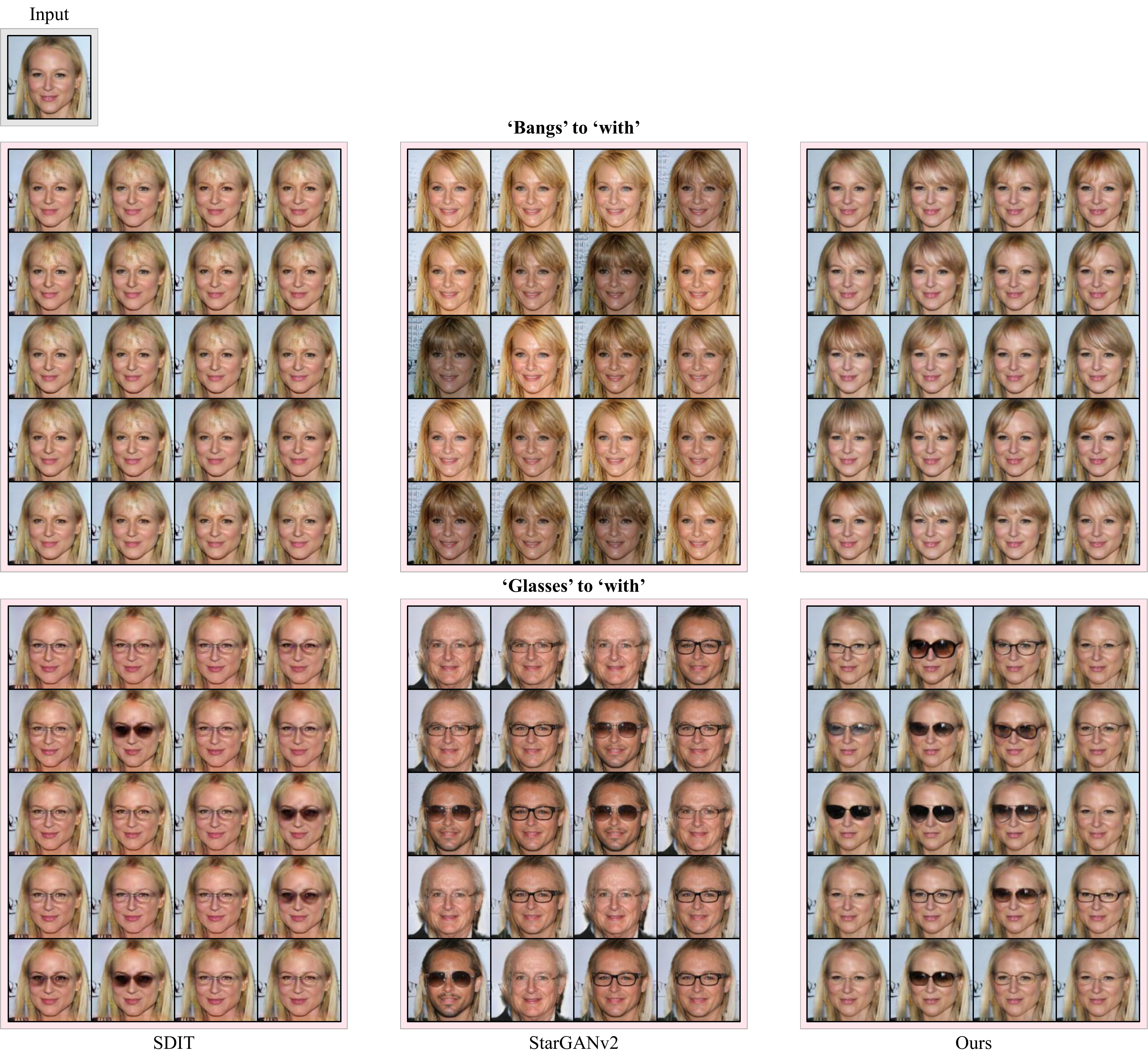}
    \caption{
     More additional qualitative results without cherry-picking for the latent-guided multi-style task.
    }
    \label{sm.2-2}
\end{figure*}

\begin{figure*}[ht]
    \centering
    \includegraphics[width=1\linewidth]{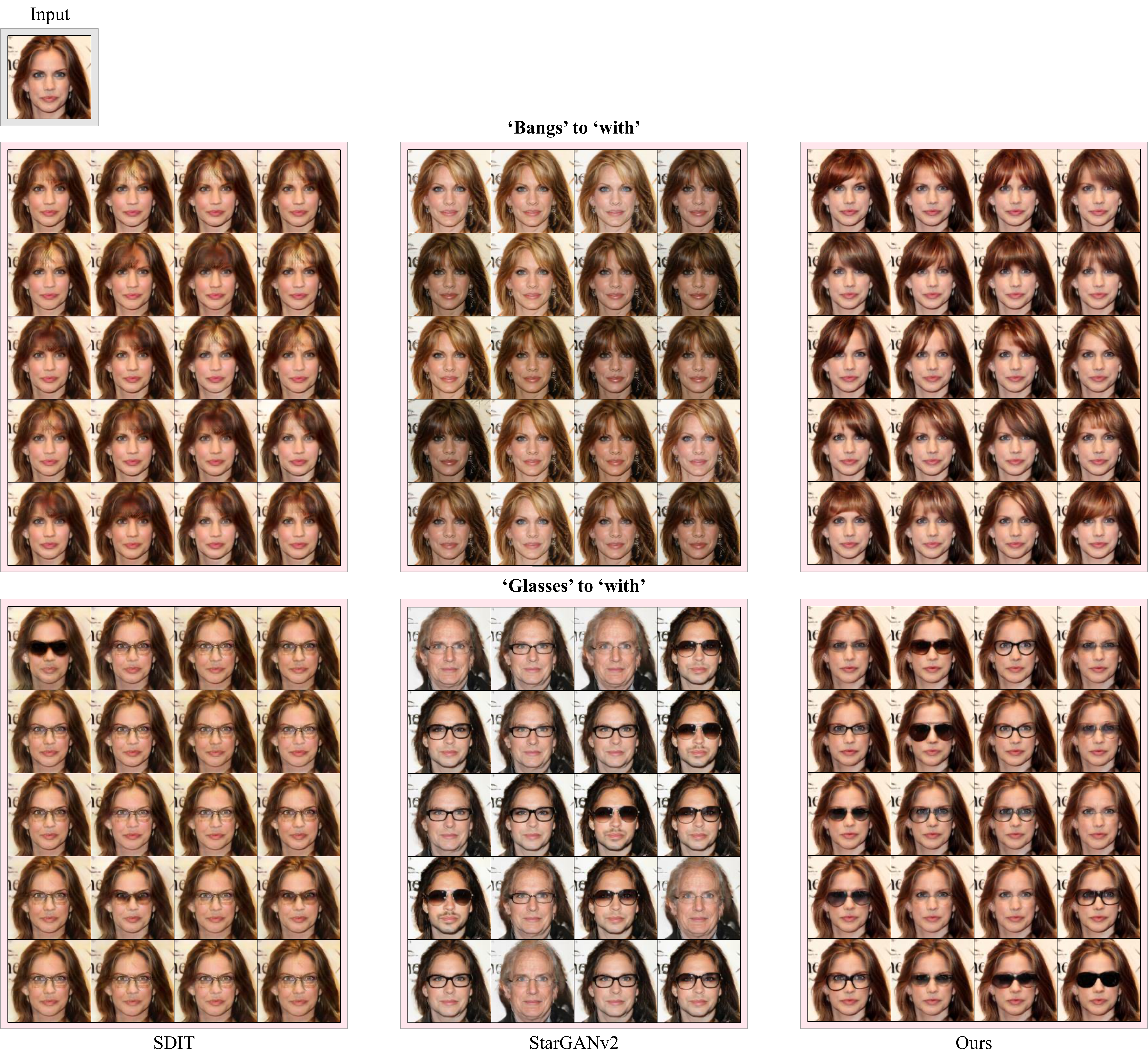}
    \caption{
     More additional qualitative results without cherry-picking for the latent-guided multi-style task.
    }
    \label{sm.2-3}
\end{figure*}

We provide some additional qualitative results without cherry-picking of baselines and our method for the latent-guided multi-style task in Figures~\ref{sm.2-1}, \ref{sm.2-2}, and \ref{sm.2-3}.
The results are completely random without manual selection.
Besides the limitations we mentioned in the paper, the baselines (\ie SDIT and StarGANv2) are observed to suffer from mode-collapse.
Furthermore, the satisfying diversity of our generated results demonstrates the effectiveness of HiSD.

\section{Interpolation of Tag-relevant Styles}

\begin{figure*}[ht]
    \centering
    \includegraphics[width=1\linewidth]{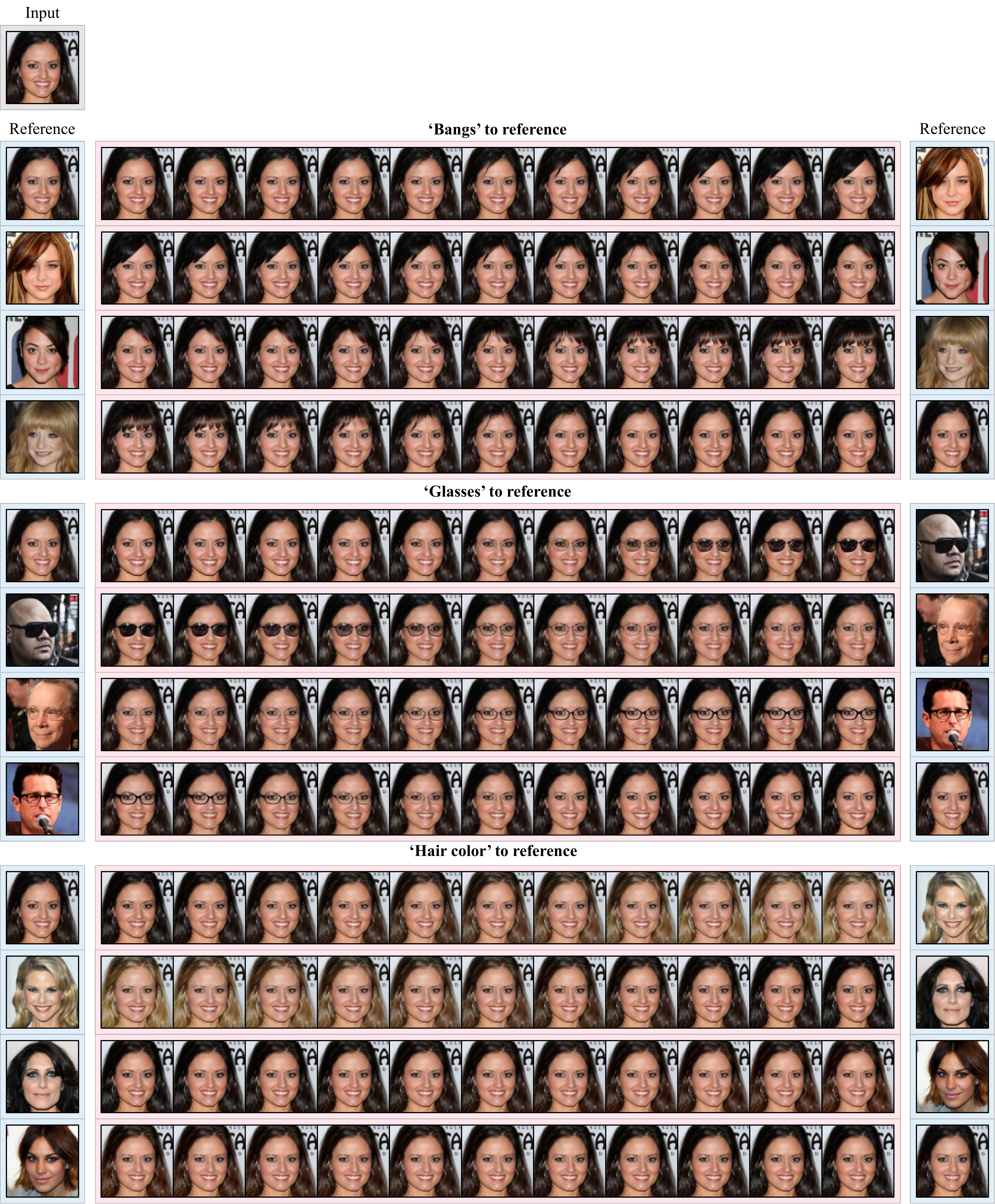}
    \caption{
     Interpolation results between two extracted tag-relevant styles for different tags. We use the linear interpolation between the style codes extracted from two different reference images to observe the continuous manipulations of the output images.
    }
    \label{sm.3-1}
\end{figure*}

\begin{figure*}[ht]
    \centering
    \includegraphics[width=1\linewidth]{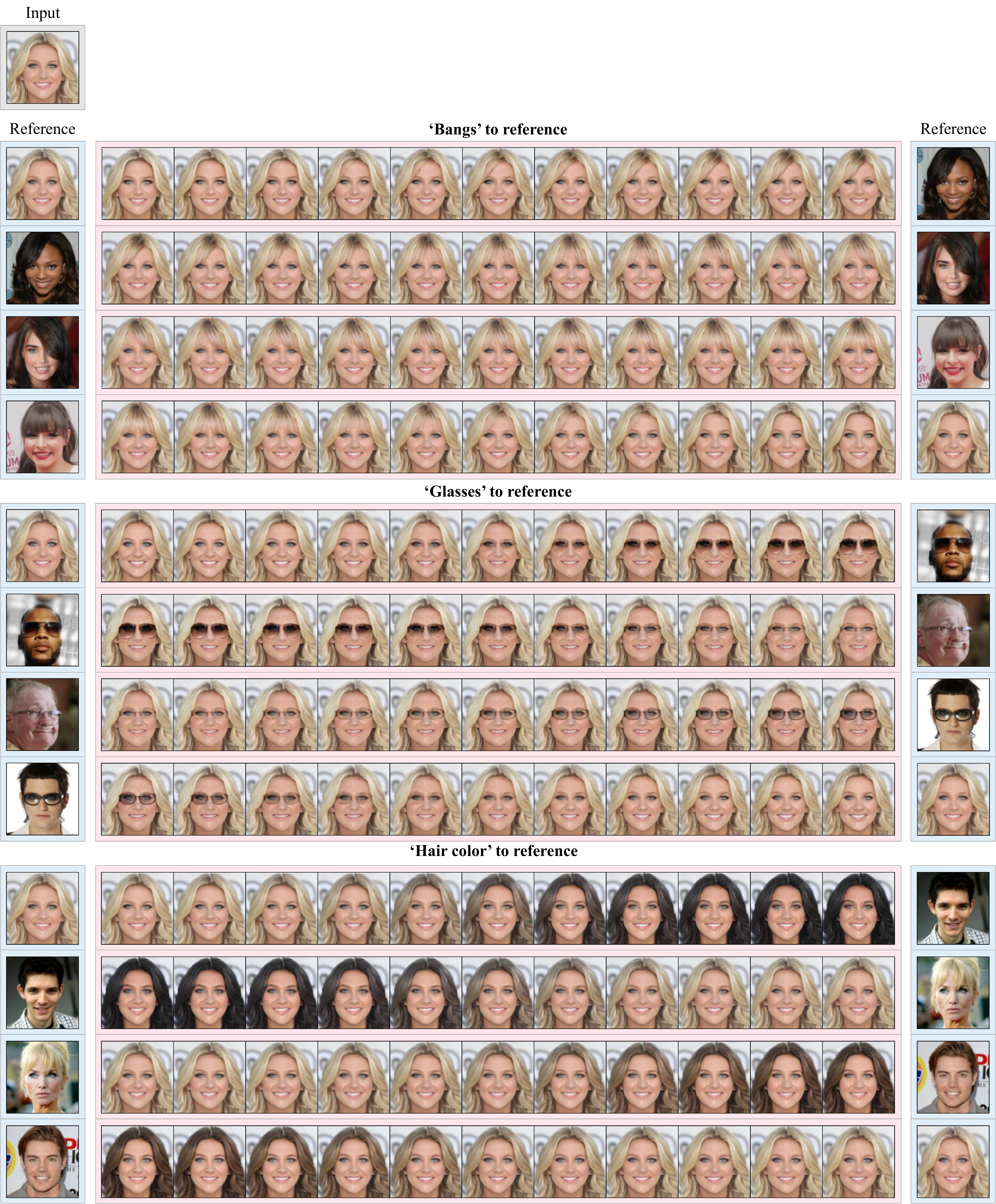}
    \caption{
     More interpolation results between two extracted tag-relevant styles for different tags. 
    }
    \label{sm.3-2}
\end{figure*}

\begin{figure*}[ht]
    \centering
    \includegraphics[width=1\linewidth]{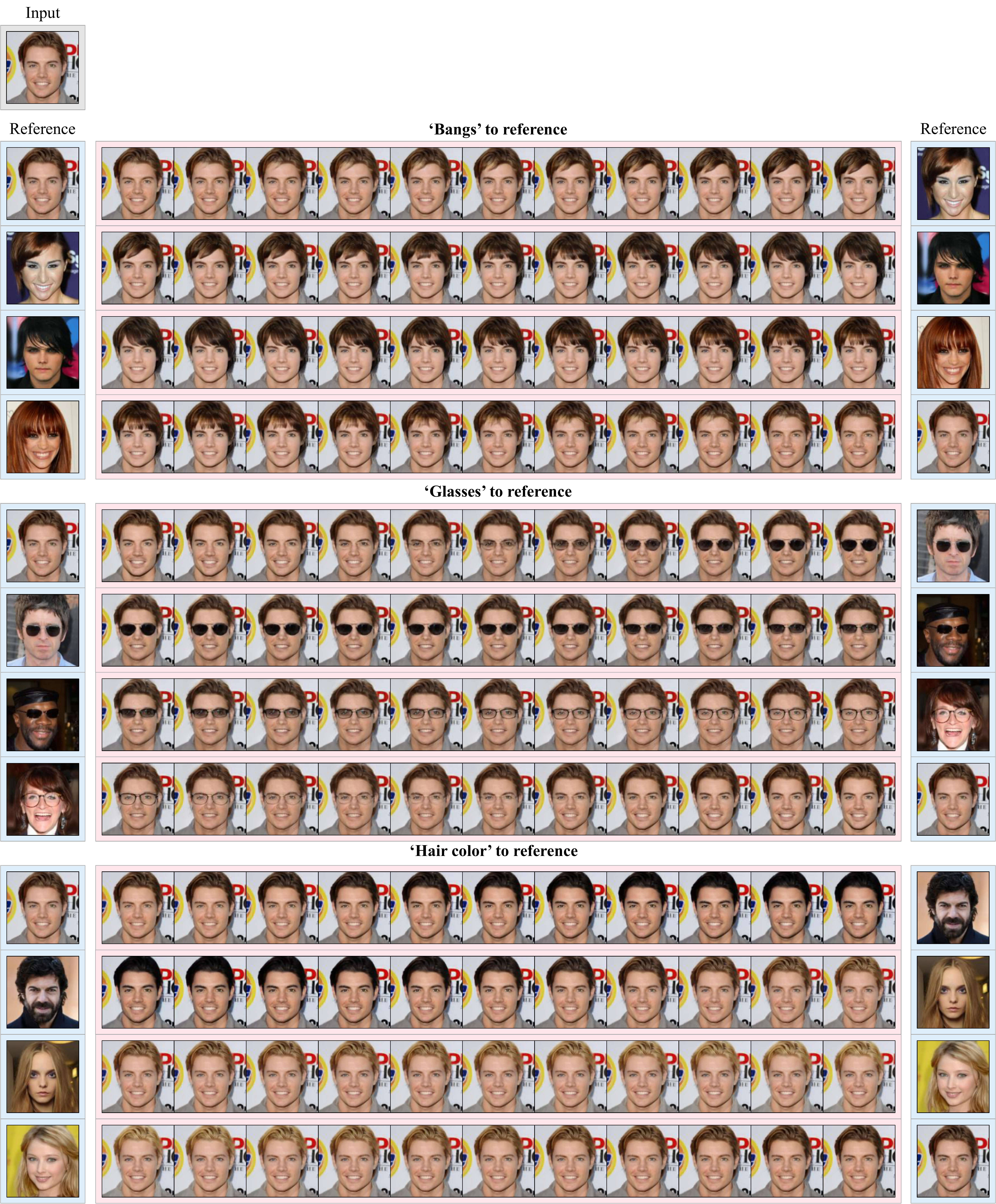}
    \caption{
     More interpolation results between two extracted tag-relevant styles for different tags. 
    }
    \label{sm.3-3}
\end{figure*}

We show the interpolation results by interpolating between the extracted tag-relevant style codes from two different reference images (with the same attribute or not) in Figures~\ref{sm.3-1}, \ref{sm.3-2}, and \ref{sm.3-3}.
The interpolation is smooth, which implies that the space of each tag-relevant style is continuous. 
The continuous tag-relevant style space allows the translations to manipulate images with a novel tag-relevant style which is not seen by the framework during training.

\section{Visualization of Tag-relevant Styles}

\begin{figure*}[ht]
    \centering
    \includegraphics[width=0.9\linewidth]{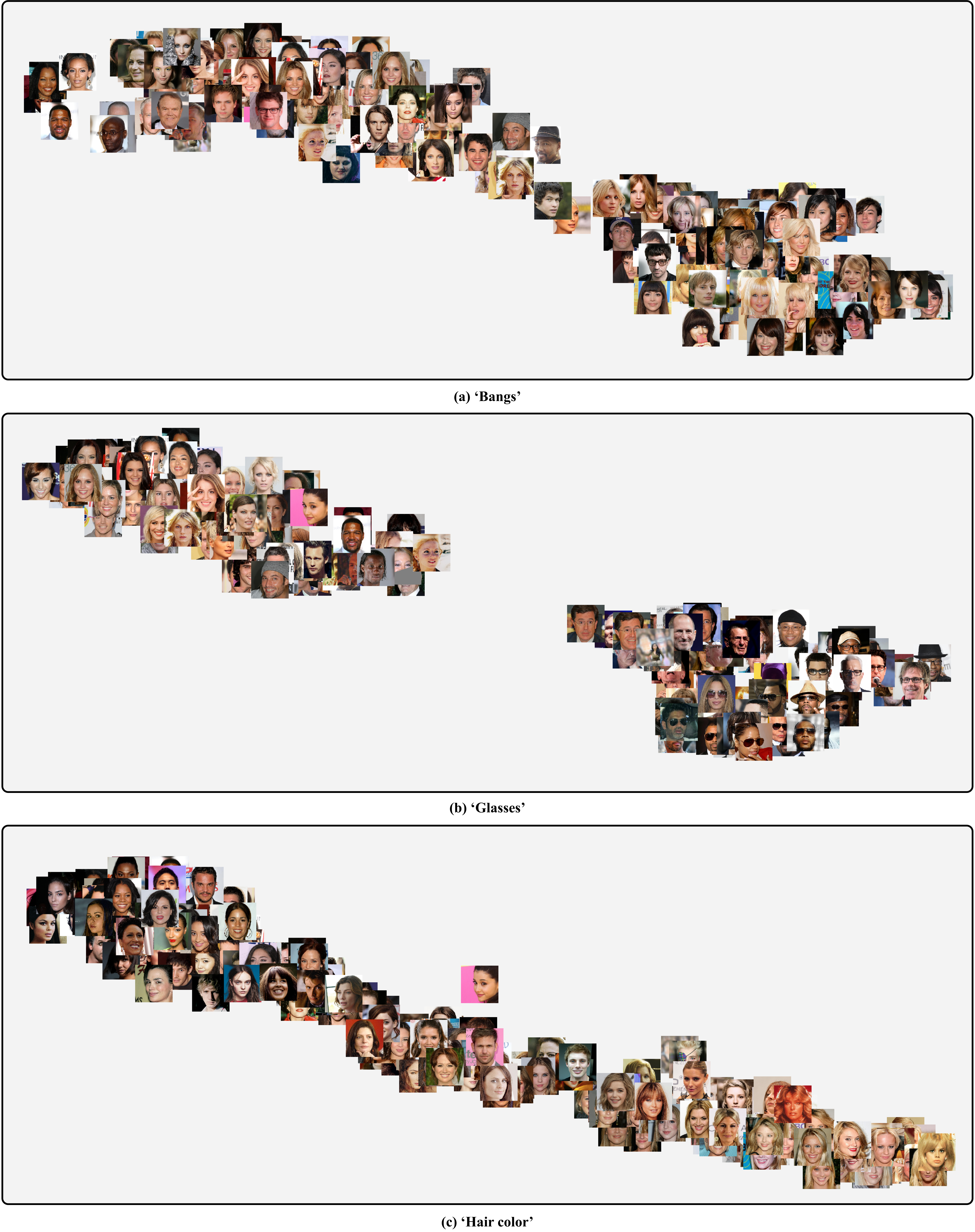}
    \caption{
     2-D representation of the extracted tag-relevant styles from 180 images using t-SNE for different tags. Please zoom-in for details.
    }
    \label{sm.4}
\end{figure*}

We further explore the style space learned by the extractor by using t-SNE to visualize the extracted tag-relevant styles in
a two-dimensional space.
As shown in Figure~\ref{sm.4}, for all tags, images with the same tag-specific attribute are grouped together in the style space.
Notably, the attribute is not inputted into the extractor in our method.
For each tag, there is a main direction and various secondary directions between different attributes.
InterFaceGAN~\cite{shen2020interfacegan} takes advantage of the main direction to manipulate attributes with unsupervised GANs~\cite{karras2019style,karras2018progressive}. 
However, it cannot guarantee the disentanglement, especially for unnecessary global and identity manipulations.
The images at the edge of the style space are always the most obvious examples for a specific attribute, while the images in the middle space are always confusing ones.
More interestingly, for tag `Hair color', the tag-relevant styles of images with attribute `brown' are extracted to be a middle state between `blond' and `black'.

\clearpage

\clearpage
{\small
\bibliographystyle{ieee_fullname}
\bibliography{egbib}
}

\end{document}